\documentclass[letterpaper, 10 pt, conference]{ieeeconf}  

\pdfoutput = 1

\UseRawInputEncoding

\IEEEoverridecommandlockouts                              

\usepackage{color}

\usepackage{amsmath} 
\usepackage{amssymb}  
\usepackage{cite}
\usepackage{footmisc}

\usepackage{algorithm}  
\usepackage{algpseudocode}  

\usepackage{booktabs}
\usepackage{makecell}
\usepackage{graphicx}
\usepackage{threeparttable}

\usepackage{cuted}

\newcommand{\bs}[1]{\boldsymbol{#1}}
\newcommand{\bsz}[1]{\boldsymbol{\rm #1}}

\newcommand{\cross}[1]{{#1}^{\times}}
\newcommand{\gravity}{{^w}\bs{g}}
\newcommand{\erot}[1]{\hat{\bsz{R}}_{i,#1}^w}
\newcommand{\epos}[1]{{^w}\hat{\bs{p}}_{i,#1}}
\newcommand{\evel}[1]{{^w}\hat{\bs{v}}_{i,#1}}
\newcommand{\dt}{\Delta t_k}

\newtheorem{definition}{Definition}
\newtheorem{remark}{Remark}
\newtheorem{proposition}{Proposition}

\usepackage{xcolor}
\usepackage{xpatch}

\makeatletter
\ExplSyntaxOn
\cs_new:Npn \bibColoredItems #1#2
  {
    \clist_map_inline:nn {#2} { \cs_new:cpn {bib@colored@##1} {#1} } 
  }
\ExplSyntaxOff

\newcommand\bib@setcolor[1]{%
  \ifcsname bib@colored@#1\endcsname
    \expandafter\color\expandafter{\csname bib@colored@#1\endcsname}
  \else
    \normalcolor
  \fi
}

\xpatchcmd\@bibitem
  {\item}
  {\bib@setcolor{#1}\item}
  {}{\fail}

\xpatchcmd\@lbibitem
  {\item}
  {\bib@setcolor{#2}\item}
  {}{\fail}
\makeatother

\title{\LARGE \bf
InGVIO: A Consistent Invariant Filter for Fast and High-Accuracy GNSS-Visual-Inertial Odometry 
}

\author{Changwu Liu$^{1}$, Chen Jiang$^{2}$ and Haowen Wang$^{3}$
\thanks{$^{1}$Changwu Liu is with School of Aerospace Engineering, Tsinghua University, Beijing, China
        {\tt\small lcw18@mails.tsinghua.edu.cn}}%
\thanks{$^{2}$Chen Jiang is with School of Aerospace Engineering, Tsinghua University, Beijing, China
        {\tt\small jc2017@mail.tsinghua.edu.cn}}%
\thanks{$^{3}$Haowen Wang is with School of Aerospace Engineering, Tsinghua University, Beijing, China
        {\tt\small bobwang@tsinghua.edu.cn}}%
}

\begin{document}

\vspace*{\fill}
\begin{strip}
\copyright{ 2023 IEEE.} Personal use of this material is permitted. Permission from IEEE must be obtained for all other uses, in any current or future media, including reprinting/republishing this material for advertising or promotional purposes, creating new collective works, for resale or redistribution to servers or lists, or reuse of any copyrighted component of this work in other works.

This is the Author Accepted Version of: C. Liu, C. Jiang and H. Wang, ``InGVIO: A Consistent Invariant Filter for Fast and High-Accuracy GNSS-Visual-Inertial Odometry,'' \emph{IEEE Robotics and Automation Letters}, 2023.

DOI: 10.1109/LRA.2023.3243520

\end{strip}
\vspace*{\fill}
\newpage

\maketitle
\thispagestyle{empty}
\pagestyle{empty}

\begin{abstract}

Combining Global Navigation Satellite System (GNSS) with visual and inertial sensors can give smooth pose estimation without drifting. The fusion system gradually degrades to Visual-Inertial Odometry (VIO) with the number of satellites decreasing, which guarantees robust global navigation in GNSS unfriendly environments. In this letter, we propose an open-sourced invariant filter-based platform, InGVIO, to tightly fuse monocular/stereo visual-inertial measurements, along with raw data from GNSS. InGVIO gives highly competitive results in terms of computational load compared to current graph-based algorithms, meanwhile possessing the same or even better level of accuracy. Thanks to our proposed marginalization strategies, the baseline for triangulation is large although only a few cloned poses are kept. Moreover, we define the infinitesimal symmetries of the system and exploit the various structures of its symmetry group, being different from the total symmetries of the VIO case, which elegantly gives results for the pattern of degenerate motions and the structure of unobservable subspaces. We prove that the properly-chosen invariant error is still compatible with all possible symmetry group structures of InGVIO and has intrinsic consistency properties. Besides, InGVIO has strictly linear error propagation without linearization error. InGVIO is tested on both open datasets and our proposed fixed-wing datasets with variable levels of difficulty and various numbers of satellites. The latter datasets, to the best of our knowledge, are the first datasets open-sourced to the community on a fixed-wing aircraft with raw GNSS.

\end{abstract}


\section{Introduction and Related Work}

\PARstart{T}{ons} of efforts in improving the performance of various filter-based Visual-Inertial Odometry (VIO) frameworks \cite{MSCKF1.0, FEJ1.0, FEJ2.0, OC-VINS, ROVIO, MSCKF-Hover, Robocentric-VIO, Exploiting-Symmetries-SLAM, OpenVINS, RI-MSCKF, Decoupled-RIEKF, EqF-VIO} in the past decade have made filter-based algorithms an efficient off-the-shelf choice for pose estimation. For a review of both optimization- and filter-based VIOs, the authors would recommend \cite{VINS-Concise-Review} instead of listing works here.

The main constraint for filter-based VIO is its inconsistency. By tackling consistency maintenance from different perspectives, modern state-of-the-art filter-based VIOs even possess competitive estimation precision compared to optimization-based methods while still keeping very low computational cost. \cite{FEJ1.0, FEJ2.0, OpenVINS} use `first estimate' values instead of their `latest best estimate' values to calculate certain Jacobians to make the filter possess the same number of unobservable directions as the true system, and thus guarantee consistency. \cite{OC-VINS} utilizes the ideal unobservable directions of VIO to adjust the matrix blocks of the state transition and the observation Jacobians through optimization, which forces the filter to behave as if it has a similar structure of the unobservable subspace like the true case. This observability constraint methodology is later used in \cite{S-MSCKF} and has been proved to improve accuracy. A few works \cite{Robocentric-VIO, ROVIO} employ robocentric frames to avoid the rank degeneration of the unobservable subspace. The above algorithms, though powerful, may consume either extra memory to store multiple copies of the same state or extra computation to transform states and adjust matrix blocks. 

Differential geometry plays vital roles in discovering the symmetries of the manifold structure of the state and the evolution of the dynamical system. These symmetries have helped develop algorithms which carry nice properties such as guaranteed convergence and consistency. \cite{InEKF-Stable-Observer} proposes the invariant EKF (InEKF), modelling state evolution on a matrix Lie group and meanwhile employing the natural matrix logarithm as nonlinear errors. InEKF has very rare local convergence properties due to its strictly linear propagation of errors and state-estimate-independent observations. \cite{InEKF-Stable-Observer} also introduces $\mathbb{SE}_2(3)$ manifold as the extended pose for SLAM (Simultaneous Localization and Mapping). Moreover, InEKF is proved to intrinsically preserve the unobservable structure of the system \cite{Consistent-EKF-SLAM,Exploiting-Symmetries-SLAM} and has been adopted within various filter-based techniques in VIO, such as the Multi-State Constraint Kalman Filter techniques (MSCKF) \cite{RI-MSCKF, CCAnalysis-RI-MSCKF} and its hybrid form with landmarks \cite{Decoupled-RIEKF}. Another approach of symmetry-preserving filters is the equivariant observer (EqF) \cite{Equi-Filter, EqF-CDC}, being also intrinsically consistent. VIO implemented by EqF propagates the system on the Lie group acting on the state manifold, as opposed to viewing the state itself as an evolving Lie group. EqF-VIO also proposes new symmetries $\mathbb{SOT}(3)$ for landmarks to make the monocular observation equation equivariant \cite{EqF-VIO, EqVIO}.

Despite efforts in VIO, four natural unobservable directions, i.e. yaw and translations, still cause the filter to drift slowly. Meanwhile, the yaw offset to the true north and the origin position of the local world frame in geographical coordinates are still unknown. By combining GNSS data, the system may become fully observable with enough number of satellites and can still output globally smooth trajectories in GNSS intermittent situation. Specifically, if the GNSS data are utilized per satellite, i.e. using raw measurements, the system can offer non-drifting estimations even if the number of satellites is below four, when GNSS alone is unable to conduct localization. Apart from triple sensor fusion using GNSS position and velocity \cite{AdaptiveFusionGVIO, LooseCoupleGVIO}, we only focus on related works using raw GNSS data in visual-inertial navigation (GVIO). \cite{GVINS, OB-GVINS, P3-GVINS} are fantastic optimization-based algorithms to fuse data from triple sensors, and meanwhile \cite{VOC-GSMSCKF, GAINS} are filter-based implementations. Though adding GNSS greatly improves the performance of the fusion system, our previous work \cite{VOC-GSMSCKF} reveals GVIO may still encounter inconsistency in various degenerate motions through observability analysis, especially when the number of satellites is not enough for independent GNSS localization. 

To the best of our knowledge, the symmetry of the fusion system involving visual, inertial and raw GNSS data is not systematically studied before. We find that GVIO only possesses `infinitesimal' symmetries as opposed to the `total' symmetries\cite{Exploiting-Symmetries-SLAM} of the VIO system. Such motion patterns constitute the degenerate motions of GVIO. Moreover, we reveal that the infinitesimal symmetry group of GVIO has various structures under different conditions. We also prove herein that invariant filter can intrinsically hold guaranteed consistency properties in variable `infinitesimal' symmetry cases, i.e. invariant error being compatible with all possible symmetry group structures. When applying MSCKF techniques in this letter, the cloned poses are not organized by a sliding window for visual updates, but are pairwise thrown in a properly-designed way to extend baseline and control complexity. Since MSCKF features are $\mathbb{SE}(3)$ instead of $\mathbb{R}^3$-type states in invariant filter implementations, they can be arbitrarily anchored to one of the $\mathbb{SO}(3)$ structures in the state formulation, e.g. \cite{RI-MSCKF}. The anchor change is coupled with the marginalization strategy, and needs careful discussions. The above describes a new visual marginalization strategy in InGVIO and the related tuning comparisons show such techniques greatly accelerate processing time with slight sacrifice of accuracy. The loss of accuracy is acceptable since in InGVIO we have one more sensor, i.e. the GNSS.

In the experiment sections, not only do we compare our InGVIO to other fusion algorithms in public datasets, but also we produce real-world data by a fixed-wing aircraft. Firstly, the number of public visual-inertial datasets with raw GNSS is far less than the number of VIO datasets. Secondly, most datasets are collected by slow-flying quadrotors, land vehicles or handheld devices. The fixed-wing datasets are themselves a contribution.

The key contributions of this letter are listed as follows:
\begin{itemize}
\item {We propose InGVIO, a symmetry-preserving invariant filter to fuse visual, inertial and raw GNSS data. InGVIO is fast due to a powerful albeit concise marginalization strategy. To the best of our knowledge, this is the first open-sourced invariant-filter-based platform\footnote{The InGVIO code and links to download our fixed-wing datasets and flight videos can be found in \textbf{https://github.com/ChangwuLiu/InGVIO}} for triple sensor fusion.}
\item {We analyze and propose GVIO has `infinitesimal' symmetry properties, being different from the VIO case. All possible structures of the GVIO symmetry group are derived. Moreover, we provide the degenerate motions of GVIO and prove InGVIO is intrinsically consistent.}
\item {We compare InGVIO on public datasets to a well-known optimization-based algorithm termed GVINS \cite{GVINS} to show its advantages among accuracy and speed.}
\item{We propose open-sourced fixed-wing datasets to validate InGVIO with various satellite numbers, which are supplements to existing datasets in the community.}
\end{itemize}

\section{Symmetries of GVIO and Degenerate Motions}\label{sec::symm_gvio}

In this section, we tackle the `infinitesimal' symmetry group structures of GVIO through symmetry group actions and derive conditions for degenerate cases. Five frames\footnote{They're the Earth-Centered-Earth-Fixed frame (ECEF), the East-North-Up frame (ENU), the world frame (w), the IMU (estimation target) frame (i) and the camera frame (c). Fig.2 of \cite{GVINS} clearly reveals their relations, and thus we omit these descriptions to save space.} are considered in GVIO estimations, and they are abbreviated in a usual way in our notation system\footnote{$\bs{p},\bs{v},\bs{b}$ herein are variables for position, velocity and biases. The left-upper corner indicates the frame of interest, and the right-bottom corner implies the object of interest, e.g. $^w\bs{p}_i$ is the IMU frame origin in the world frame. Variable with $\hat{\cdot}$ implies its estimation value, and that with nothing above implies its true value. Variables with $\delta$-symbol at the left are the error states. $\bsz{R}_\text{frame1}^\text{frame2},\bsz{T}_\text{frame1}^\text{frame2}$ are $\mathbb{SO}(3),\mathbb{SE}(3)$ transformations from `frame1' to `frame2'. Specifically, $\delta\bs{\theta}$ is chosen to represent the error state for $\mathbb{SO}(3)$.}. Unbiased GVIO system is modelled herein since gyroscope bias $\bs{b}_g$ and accelerometer bias $\bs{b}_a$ are even observable in VIO \cite{ConAnalysisVINS}. 

\subsection{Infinitesimal Symmetries of System on Lie Groups}

The motivation is that raw GNSS measurement equations possess a nonlinear norm structure which does not allow direct applications of the `total invariance' definition in VIO.

\begin{definition}
Let $\chi_t$ be a state on a Lie group $G$ with dynamical evolution $\dot{\chi}_t=f(\chi_t,u_t)$ and discrete observation $y_{t_k}=h(\chi_{t_k})$. Such system is called \textbf{infinitesimally invariant} under the left action $\rhd$ of Lie group $S$ on $G$, if $\forall s\in S$, we have $\rhd_{s*}(\dot{\chi}_t)=f(s\rhd\chi_t,u_t)$ and $\frac{\partial}{\partial s}\vert_{s=id_S}h(s\rhd\chi_{t_k})=0$, where $\rhd:S\times G\rightarrow G$ is a well-defined group action and $\rhd_{s*}:\mathcal{T}G\rightarrow\mathcal{T}G$ is the induced linear push-forward map between the tangent bundles $\mathcal{T}G$ of $G$ by a fixed $s\in S$.
\end{definition}

\begin{remark}
$\frac{\partial}{\partial s}\vert_{s=id_S}h(s\rhd\chi_{t_k}):=\frac{d}{d\lambda}\vert_{\lambda=0}h(\exp(\lambda A)\rhd\chi_{t_k})$, which means taking the derivative along a curve induced by $A\in T_eS$. $T_eS$ is the Lie algebra of the infinitesimal symmetry group $S$, and the exponential is from $T_eS$ to $S$. 
\end{remark}

\begin{remark}
Infinitesimal invariance is a \textbf{weakened version of total invariance} defined in \cite{Exploiting-Symmetries-SLAM}. The only difference is that we only require the observations remain unchanged under the $S$-group action to just first order.  
\end{remark}

The group $S$ describes the infinitesimal symmetries of the system on Lie group $G$. To use error state observers of such system in a vector space, we need to define two maps $\boxplus:G\times\mathbb{R}^n\rightarrow G$ and $\boxminus:G\times G\rightarrow\mathbb{R}^n$ to regulate the behaviors of the error, i.e. $\chi=\hat\chi\boxplus\delta\chi$ and $\delta\chi=\chi\boxminus\hat\chi$.

\begin{proposition}
For a system on Lie group $G$ with initial value $\chi_0$ infinitesimally invariant under the left action of group $S$, column vectors $N=\frac{\partial}{\partial s}\vert_{s=id_S}\left[(s\rhd\chi_0)\boxminus\chi_0\right]$ lie in the right nullspace of the observability matrix of its error state EKF evaluated by true values. Each column of $N$ is the so-called ideal unobservable direction.  
\end{proposition}

\emph{Proof:} Steps to prove Proposition 1 of \cite{Exploiting-Symmetries-SLAM} are also suitable for this infinitesimally invariant case. $\square$

\subsection{Conditional Infinitesimal Symmetries of GVIO}

Let the GVIO state be $\chi=(\bsz{T}_i^w,{^w}\bs{v}_i, \bsz{T}_{c_m}^w,{^w}\bs{p}_{f_j})$ in our notation, where $\bsz{T}_{c_m}^w$ is the $m$-th historical camera pose kept in the filter and ${^w}\bs{p}_{f_j}$ is the $j$-th landmark position. $(\bsz{T}_i^w, {^w}\bs{v}_i)$ is the pose and velocity of the IMU frame. Note that we do not involve GNSS related states such as clock biases and frequency shifts just in this section, meaning we must later modify the GNSS measurement model to cancel these terms. The process model for GVIO is shown in (\ref{gvio_proc}). 
\begin{gather}
\label{gvio_proc}
\begin{split}
\dot{\bsz{T}}_i^w=\bsz{T}_i^w\bsz{U},{^w}\dot{\bs{v}}_i=\bsz{R}_i^w\tilde{\bs{a}}+\gravity,\dot{\bsz{T}}_{c_m}^w=0,{^w}\dot{\bs{p}}_{f_j}=0
\end{split}
\end{gather}
Unbiased angular rate and acceleration are marked as $\tilde{\bs{\omega}},\tilde{\bs{a}}$, and $\bsz{U}$ is from the Lie algebra of $\mathbb{SE}(3)$ with the form $\bsz{U}=\begin{bmatrix}\cross{\tilde{\bs{\omega}}} & \bsz{R}_i^{wT}{^w}\bs{v}_i \\ 0^T & 0\end{bmatrix}$. $\times$ implies the skew-symmetric matrix of $\mathbb{R}^3$. The visual measurement related to the $j$-th feature and the $m$-th camera pose is shown in (\ref{gvio_vis_meas}).
\begin{gather}
\label{gvio_vis_meas}
\bs{r}_j^{(m)}=\pi((\bsz{T}_{c_m}^w)^{-1}{^w}\bs{p}_{f_j}), \pi:(x_1,x_2,x_3)\mapsto\begin{bmatrix}x_1/x_3 \\ x_2/x_3\end{bmatrix}
\end{gather}
Define a subgroup of $\mathbb{SE}(3)$ by a semidirect product structure $S:=\mathbb{SO}_g(2)\ltimes\mathbb{R}^3$, where $\mathbb{SO}_g(2)$ indicates rotation along the direction of $\gravity$. The $S$-action $\rhd$ on $\chi$ is defined as $\bsz{T}_S\rhd\chi\mapsto(\bsz{T}_S\bsz{T}_i^w,\bsz{R}_S{^w}\bs{v}_i, \bsz{T}_S\bsz{T}_{c_m}^w,\bsz{T}_S{^w}\bs{p}_{f_j})$, where $\bsz{R}_S$ is the $\mathbb{SO}_g(2)$ rotation part of $\bsz{T}_S\in S$. Both (\ref{gvio_proc}) and (\ref{gvio_vis_meas}) are \textbf{totally invariant} under the left action of $S$. These are the well-known unobservable directions of VIO. To simplify analysis herein, we consider only one constellation with GNSS involved, though our implementation supports four. The pseudo range $\rho^{k}$ and Doppler shift $\dot{\rho}^{k}$ (pseudo-range rate) corresponding to the $k$-th satellite are listed as (\ref{gvio_gnss_meas}).
\begin{gather}
\label{gvio_gnss_meas}
\begin{split}
\rho^{k}&=\Vert\bsz{T}_w^\text{ECEF}{^w}\bs{p}_i-\bs{p}_{sat}^{k}\Vert+c(t-\Delta t^{k})+D^k \\
\dot{\rho}^{k}&=-\bs{n}_k^{T}(\bsz{R}_w^\text{ECEF}{^w}\bs{v}_i-\bs{v}_{sat}^k)+c(f-\Delta f^k)
\end{split}
\end{gather}
Rationales of (\ref{gvio_gnss_meas}) can be found in \cite{GPSreceiver}. The rotation $\bsz{R}_w^\text{ECEF}$ and isometry $\bsz{T}_w^\text{ECEF}$ are constants provided by the initializer. $\bs{p}_{sat}^k,\bs{v}_{sat}^k$ are the position and velocity of the $k$-th satellite in ECEF. $\bs{n}_k$ is the unit vector pointing from the receiver to the $k$-th satellite. $\bs{n}_k$ varies slowly and is viewed constant in symmetry analysis. $c$ is the light speed. $t$ is the clock bias of the local system to GNSS time and $f$ is the clock bias random walk, indicating the frequency shift of the local receiver. Both $t$ and $f$ are unknown GNSS states to be estimated. $\Delta t^k$ and $\Delta f^k$ are the broadcasted clock bias and shift of the onboard satellite clock. $D^k=I_d^k+T_d^k+S^k$ is the propagation delay consisting of ionospheric \cite{IonoDelay}, tropospheric \cite{TropoDelay} and Sagnac Effect \cite{SagnacEffect} contributions. $D^k$ is considered known by prior models with broadcasted parameters. For symmetry analysis, cancelling $(t, f)$ by conducting single difference for both pseudo-ranges and Doppler shifts between the $k$-th and $l$-th satellite at the same timestamp yields (\ref{gvio_gnss_diff_meas}).
\begin{gather}
\label{gvio_gnss_diff_meas}
\begin{split}
  \tilde\rho^{kl}&=\Vert\bsz{T}_w^\text{ECEF}{^w}\bs{p}_i-\bs{p}_{sat}^{k}\Vert-\Vert\bsz{T}_w^\text{ECEF}{^w}\bs{p}_i-\bs{p}_{sat}^{l}\Vert \\
  \tilde{\dot{\rho}}^{kl}&=-\bs{n}_k^{T}\bsz{R}_w^\text{ECEF}{^w}\bs{v}_i+\bs{n}_l^{T}\bsz{R}_w^\text{ECEF}{^w}\bs{v}_i
\end{split}
\end{gather}
Please note the differences of pseudo range $\tilde{\rho}^{kl}$ and Doppler shift $\tilde{\dot\rho}^{kl}$ in ($\ref{gvio_gnss_diff_meas}$) also absorb the remaining constants in (\ref{gvio_gnss_meas}). (\ref{gvio_gnss_diff_meas}) is compatible with $\chi$ without any unknown GNSS states.

The norm and dot-product structure in (\ref{gvio_gnss_diff_meas}) make it impossible to hold the total invariance of $S$-action. However, infinitesimal invariance may exist under certain circumstances.

\begin{proposition}
Consider the number of satellites being $N$. Stack rows $(\bs{n}_j-\bs{n}_{j-1})^T\bsz{R}_w^\text{ECEF},j=2...N$ to define matrix $\bsz{N}_g$. If $\text{rank}(\text{null}(\bsz{N}_g))>0$, GVIO has \textbf{infinitesimal symmetries}, i.e. infinitesimally invariant under the action of $H$. Moreover, if $\gravity^\times{^w}\bs{p}_i\in\text{null}(\bsz{N}_g)$ and $\gravity^\times{^w}\bs{v}_i\in\text{null}(\bsz{N}_g)$, then $H=\mathbb{SO}_g(2)\ltimes\text{null}(\bsz{N}_g)$. Otherwise, $H=\text{null}(\bsz{N}_g)$. 
\end{proposition}

\emph{Proof:} $H$ is a subgroup of $S$, and thus the $H$-action on $\chi$ inherits that of $S$. Besides, total $H$-invariance of (\ref{gvio_proc}) and (\ref{gvio_vis_meas}) naturally holds, and thus is infinitesimally invariant. Denote $\bs{t}_g\in\text{null}(\bsz{N}_g)$, and substitute ${^w}\bs{p}_i\mapsto{^w}\bs{p}_i+\lambda \bs{t}_g$ in (\ref{gvio_gnss_diff_meas}). We have $\frac{d}{d\lambda}\vert_{\lambda=0}\tilde\rho^{kl}=0$ and $\frac{d}{d\lambda}\vert_{\lambda=0}\tilde{\dot\rho}^{kl}=0$ (*). However, if we substitute ${^w}\bs{p}_i\mapsto\exp(\lambda\gravity^\times){^w}\bs{p}_i$ and ${^w}\bs{v}_i\mapsto\exp(\lambda\gravity^\times){^w}\bs{v}_i$ in (\ref{gvio_gnss_diff_meas}), only with conditions $\gravity^\times{^w}\bs{p}_i,\gravity^\times{^w}\bs{v}_i\in\text{null}(\bsz{N}_g)$ can we establish (*). $\square$ 

\begin{remark}
When the number of satellites is lower than 4, degenerate motion happens. Certain translations determined by satellite geometric structure $\bsz{N}_g$ yield the unobservable translations of GVIO. The yaw of GVIO can be unobservable when the motion pattern follows conditions in proposition 2. 
\end{remark}

\begin{remark}
Partial unobservability of GVIO indicates inconsistency happening, which will be discussed later. The elegant and concise analysis using infinitesimal group action gives equivalent results compared to our previous work \cite{VOC-GSMSCKF} using total observability matrix. Following proposition 1 will give the exact same basis of its right nullspace.  
\end{remark}

\section{InGVIO Filter Framework}\label{sec::filter_framework}

\subsection{State on Matrix Lie Group and Its Error Form}

Let the state of InGVIO be $\chi = (\bsz{R}_i^w, {^w}\bs{p}_i, {^w}\bs{v}_i, \bs{b}_g, \bs{b}_a, \\ \bsz{R}_c^i, {^i}\bs{p}_c, \bsz{R}_{c_m}^w, {^w}\bs{p}_{c_m},  t_\alpha, f)$, and $\chi$ evolves on a product structure $\mathbb{SE}_2(3)\times(\mathbb{R}^3)^2\times\mathbb{SE}(3)\times(\mathbb{SE}(3))^N\times\mathbb{R}^4\times\mathbb{R}$. $(\bsz{R}_i^w,{^w}\bs{p}_i,{^w}\bs{v}_i)\in\mathbb{SE}_2(3)$ is the IMU extended pose and $(\bs{b}_g,\bs{b}_a)\in(\mathbb{R}^3)^2$ are the gyroscope and accelerometer biases. $(\bsz{R}_c^i,{^i}\bs{p}_c)$ is the (left) camera extrinsics. $(\bsz{R}_{c_m}^w,{^w}\bs{p}_{c_m})$ is one of the $N$ historical poses of the camera frame. $t_\alpha$ is the clock bias of the local system to one of the four constellations, and $f$ is the random-walk of the clock biases. With $\Gamma$-functions (see Appendix), the $\mathbb{R}^{26+6N}$ error corresponding to $\chi$ is defined in (\ref{ingvio_boxplus}). This is the right-invariant matrix logarithmic error of the above state manifold structure.
\begin{gather}
\label{ingvio_boxplus}
\begin{split}
&(\hat{\bsz{R}}_i^w, {^w}\hat{\bs{p}}_i, {^w}\hat{\bs{v}}_i)\boxplus(\delta\bs{\theta}_i^w, \delta{^w}\bs{p}_i, \delta{^w}\bs{v}_i)=\\ &(\Gamma_0(\delta\bs{\theta}_i^w)\hat{\bsz{R}}_i^w, \Gamma_0(\delta\bs{\theta}_i^w){^w}\hat{\bs{p}}_i+\Gamma_1(\delta\bs{\theta}_i^w)\delta{^w}\bs{p}_i, \\ &\Gamma_0(\delta\bs{\theta}_i^w){^w}\hat{\bs{v}}_i+\Gamma_1(\delta\bs{\theta}_i^w)\delta{^w}\bs{v}_i) \\ &(\hat{\bs{b}}_g, \hat{\bs{b}}_a)\boxplus( \delta\bs{b}_g, \delta\bs{b}_a)=(\hat{\bs{b}}_g+\delta\bs{b}_g, \hat{\bs{b}}_a+\delta\bs{b}_a) \\ &(\hat{\bsz{R}}_c^i, {^i}\hat{\bs{p}}_c)\boxplus(\delta\bs{\theta}_c^i, \delta{^i}\bs{p}_c)=(\Gamma_0(\delta\bs{\theta}_c^i)\hat{\bsz{R}_c^i}, \\ &\Gamma_0(\delta\bs{\theta}_c^i){^i}\hat{\bs{p}}_c+\Gamma_1(\delta\bs{\theta}_c^i)\delta{^i}\bs{p}_c) \\ &(\hat{\bsz{R}}_{c_m}^w, {^w}\hat{\bs{p}}_{c_m})\boxplus(\delta\bs{\theta}_{c_m}^w, \delta{^w}\bs{p}_{c_m})=(\Gamma_0(\delta\bs{\theta}_{c_m}^w)\hat{\bsz{R}_{c_m}^w}, \\ &\Gamma_0(\delta\bs{\theta}_{c_m}^w){^w}\hat{\bs{p}}_{c_m}+\Gamma_1(\delta\bs{\theta}_{c_m}^w)\delta{^w}\bs{p}_{c_m}) \\
&(\hat{t}_\alpha, \hat{f})\boxplus(\delta t_\alpha, \delta f)=(\hat{t}_\alpha+\delta t_\alpha, \hat f+\delta f)
\end{split}
\end{gather} 

\subsection{Intrinsic Consistency Property of InGVIO}

Unlike FEJ\cite{FEJ1.0, FEJ2.0}, invariant filter-based algorithm does not rely on the `disturbance term' of the estimated observability matrix compared to the ideal case, and thus such term is omitted here. Readers can refer to our previous \cite{VOC-GSMSCKF} for the proof of conditional inconsistency in GVIO. 

\begin{definition}
The error $\boxminus$ is said to be \textbf{compatible with the infinitesimal symmetry group} $H$ of GVIO if $\frac{\partial}{\partial h}((h\rhd\chi)\boxminus\chi)\vert_{h=id_H}$ does not depend on $\chi$. This compatibility criterion guarantees intrinsic consistency of the error-state observer even in the infinitesimally invariant case.
\end{definition}

\begin{remark}
A key observation is that the proof of theorem 1 of \cite{Exploiting-Symmetries-SLAM} only relies on `Jacobians' and thus covers not only totally but also infinitesimally invariant case.
\end{remark}

\begin{proposition}
The error state defined by (\ref{ingvio_boxplus}) is compatible with all possible structures of $H$ listed in Proposition 2. 
\end{proposition}

\emph{Proof:} Direct computation of $\frac{\partial}{\partial h}\vert_{h=id}((h\rhd\hat{\chi})\boxminus\hat{\chi})$ for $H=\mathbb{SO}_g(2)\ltimes\text{null}(\bsz{N}_g)$ or $\text{null}(\bsz{N}_g)$ yields results independent of $\hat\chi$, and thus InGVIO is consistent. $\square$ 

\subsection{Propagation and Augmentation}

The InGVIO state is propagated analytically from $t_k$ to $t_{k+1}$ by the assumption that the IMU measurements hold unchanged between consecutive time-steps, as shown in (\ref{ingvio_dis_prop}).
\begin{gather} 
\label{ingvio_dis_prop}
\begin{split}
\erot{k+1}=&\erot{k}\Gamma_0(\tilde{\bs{\omega}}_k\dt) \\
\evel{k+1}=&\evel{k}+\gravity\dt+\erot{k}\Gamma_1(\tilde{\bs{\omega}}_k\dt)\tilde{\bs{a}}_k\dt \\
\epos{k+1}=&\epos{k}+\evel{k}\dt+\frac{1}{2}\gravity\dt^2 \\
&+\erot{k}\Gamma_2(\tilde{\bs{\omega}}_k\dt)\tilde{\bs{a}}_k\dt^2 \\
\hat{t}_{\alpha,k+1}=&\hat{t}_{\alpha,k}+\hat{f}_k\dt (\alpha=\text{GPS, BDS, GAL, GLO})
\end{split}
\end{gather}
The time interval is $\Delta t_k:=t_{k+1}-t_k$. For raw IMU measurements $(\bs{\omega}_k,\bs{a}_k)$ at $t_k$, $\tilde{\bs{\omega}}_k=\bs{\omega}_k-\hat{\bs{b}}_{g,k}$ and $\tilde{\bs{a}}=\bs{a}_k-\hat{\bs{b}}_{a,k}$ are unbiased IMU gyroscope and accelerometer inputs. All other states not appearing in (\ref{ingvio_dis_prop}) remain the same during propagation. The discrete-time state transition matrix from $t_k$ to $t_{k+1}$ for the error state defined in (\ref{ingvio_boxplus}) has a block diagonal structure\footnote{$\bsz{I}_n$ or $\bsz{0}_n$ denotes an $n\times n$ identity or zero matrix.} $\bsz{\Phi}_{k}=\text{diag}(\bsz{\Phi}_\text{IMU}, \bsz{I}_6, \bsz{I}_{6N}, \bsz{\Phi}_\text{GNSS})$. $\bsz{\Phi}_\text{IMU}\in\mathbb{R}^{15\times 15}$ can also be analytically solved using $\Gamma$-functions and such expression is the same as the right-invariant case in \cite{ContactAid}. $\bsz{\Phi}_\text{GNSS}\in\mathbb{R}^{5\times 5}$ is trivially acquired by the matrix exponential of a nilpotent matrix to just first order. $\bsz{\Phi}_k$ is independent of the estimated states due to the state-independent and strictly linear propagation \cite{InEKF-Stable-Observer} of the error states in the invariant filter, preserving the Gaussian distribution of such error. With $\bsz{\Phi}_k$, the covariance of the error state can be propagated with IMU measurement noises and the random walk noises of their biases. InGVIO does not take the hybrid form to simultaneously keep landmark states ${^w}\bs{p}_{f_j}$ due to an efficient trade-off in visual updates. The camera pose in monocular configuration or just the left camera pose in stereo case is statistically augmented \cite{MSCKF1.0} to the state after propagation, and these historical poses kept in the state are termed `cloned poses' in MSCKF techniques. The augmentation steps of InGVIO resemble those in \cite{S-MSCKF, OpenVINS} and thus are omitted here.

\subsection{Marginalization Strategy and Anchor Change}

\begin{table}[t]
  \centering
  \caption{Accuracy and Average Time per Frame Comparison on Various Visual Marginalization Strategies in Invariant VIO}
  \begin{tabular}{ccccccccc}
  \toprule
  KF/SW & Num-Lmks & Max-Clones & RMSE (m) & Time (ms) \\
  \midrule
  KF & 0 & 35 & 4.50/4.07 & 3.97/4.29 \\
  \textbf{KF} & \textbf{0} & \textbf{30} & \textbf{4.58/4.13} & \textbf{2.31/3.49} \\
  KF & 0 & 25 & 5.35/4.97 & 1.75/2.86 \\
  SW & 0 & 35 & 4.55/4.30 & 5.52/10.86 \\
  SW & 20 & 35 & 4.31/4.27 & 10.78/16.24 \\
  SW & 0 & 30 & 4.63/4.36 & 4.03/8.39 \\
  SW & 20 & 30 & 4.54/4.35 & 8.49/12.86 \\
  SW & 0 & 25 & 5.72/5.28 & 2.73/6.01 \\
  SW & 20 & 25 & 5.34/5.11 & 5.65/9.89 \\ 
  \bottomrule    
  \end{tabular}
  \begin{tablenotes}
      \footnotesize
      \item Tests of pure invariant VIO are conducted on a PC with Intel i7-12700 @ 32 GB memory in simulation environments for around 400 meters. Statistics involve `monocular/stereo' configurations. All other conditions are kept the same (max points per frame = 150, IMU and visual noises). `Num-Lmks' implies the number of landmarks added into the state. `KF' is the strategy proposed herein, and `SW' indicates sliding window marginalizing only the oldest clone. The bold gives the best trade-off. 
  \end{tablenotes}
  \label{tab::compare_marg}
\end{table}

\begin{figure}[t]
    \centering
    \parbox{3in}{
        \centering
    \includegraphics[scale=0.40]{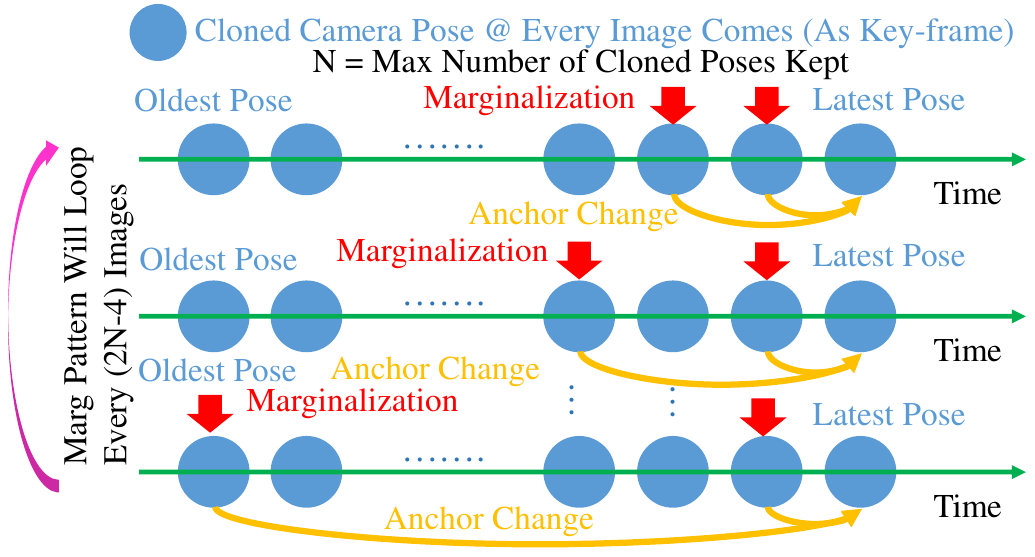}
    }
    \caption{Marginalization strategies when exceeding preset max number of cloned poses. After removal of lost-tracked features, those MSCKF features involved in the two target frames are used to construct visual updates. The features anchored to those frames are switched to the newest clone.}
    \label{fig::keyframe_select}
\end{figure}

\begin{algorithm}[t]  
  \caption{Monocular/Stereo Visual Updates in InGVIO}  
  \label{alg::ingvio_vis_update}  
  \begin{algorithmic}[1]  
    \Repeat { Every new image comes. (treated as key-frame)}
      \State Remove lost-tracked features and do MSCKF update.
      \State Select two poses for marginalization as Fig. \ref{fig::keyframe_select}, and update using features involved in these two poses.
      \State Change anchor for features anchored at those poses.
      \State Remove the two selected poses from InGVIO state. 
    \Until{Navigation stops.}  
  \end{algorithmic}  
\end{algorithm}

The sliding window approach saves computational load and utilizes MSCKF updates in the recent work on invariant VIO \cite{Decoupled-RIEKF} compared to earlier works \cite{Consistent-EKF-SLAM,Exploiting-Symmetries-SLAM} which directly add landmark positions in the state. \cite{Decoupled-RIEKF} is based on \cite{OpenVINS} and most publicly-available parameter tunings of the latter project maintain a sliding window with time-duration around 1 second. To improve accuracy, \cite{OpenVINS} also puts lots of SLAM features tracked longer than the sliding window in the state. However, the anchor change of SLAM features are much more complex than MSCKF features because the former involves computing and substituting rows and columns in the total covariance matrix. In contrast to \cite{OpenVINS,Decoupled-RIEKF} marginalizing only the oldest clone, we propose a marginalization strategy in Fig. \ref{fig::keyframe_select}, where we recurrently choose two clones to marginalize in every alternate image. Every image is viewed as a key-frame to guarantee the frequency for visual updates is at least half of that of the image. To avoid triangulation failure in stationary motions, the features corresponding to the second-latest frame and one of the older frames are chosen to construct updates. The advantages for our strategy are: (a) extending the longest baseline to almost twice compared to a sliding window method \cite{OpenVINS,Decoupled-RIEKF} with the same number of cloned poses kept; (b) compressing the dimension of sub-states corresponding to features contained in the two target poses compared to algorithms marginalizing more than two poses at once; (c) cancelling SLAM-type features to reduce the dimension of the state and saving computational expenses for anchor change; (d) automatically keeping relatively old poses without extra computation for selection compared to \cite{MSCKF-Hover}. It's clear that our key-frame selection algorithm is a trade-off between accuracy and complexity in Table \ref{tab::compare_marg}. Though slight loss of accuracy happens, our proposed strategy greatly accelerates the visual fusion. A simplification in visual updates is acceptable because GNSS is involved, and we do not require the system runs in pure VIO mode for a long period. The sequence of our strategy is in Alg. \ref{alg::ingvio_vis_update}. The visual Jacobians corresponding to the $j$-th feature observed by the $m$-th camera pose are shown in (\ref{ingvio_visual_jacobian}) using (\ref{ingvio_boxplus}).
\begin{gather}
\label{ingvio_visual_jacobian}
\begin{split}
\delta{^{c_m}}\bs{p}_{f_j}=\space\space&\hat{\bsz{R}}_{c_m}^{wT}{^w}\hat{\bs{p}}_{f_j}^\times\delta\bs{\theta}_{c_m}^w-\hat{\bsz{R}}_{c_m}^{wT}{^w}\hat{\bs{p}}_{f_j}^\times\delta\bs{\theta}_{c_0}^w \\
&-\hat{\bsz{R}}_{c_m}^{wT}\delta{^w}\bs{p}_{c_m}+\hat{\bsz{R}}_{c_m}^{wT}\delta{^w}\bs{p}_{f_j}
\end{split}
\end{gather} 

${^{c_m}}\bs{p}_{f_j}$ is the position of the $j$-th feature in the $m$-th camera frame (left camera for the stereo case, assuming $\mathbb{SE}(3)$ from the left to right camera is precisely calibrated) and $c_0$ denotes the anchored pose. The rest techniques including nullspace projection and QR decomposition to compress rows resemble traditional MSCKF \cite{MSCKF1.0} and thus are omitted. 

\subsection{GNSS Initialization and Update}

Most off-the-shelf camera-IMU kits do not support external triggering, and thus no hardware synchronization is guaranteed between the GNSS module and the camera system. Each GNSS message contains a header with GNSS timestamp, which can be converted to the UTC time with pre-aligned constant time offset. Soft synchronization is conducted in InGVIO to find the nearest GNSS measurements to every image. In practice, we only use pseudo range and Doppler shift measurements which suit this soft synchronization policy quite well.   

The $\mathbb{SE}(3)$ transformation $\bsz{T}_w^\text{ECEF}=\bsz{T}_\text{ENU}^\text{ECEF}\bsz{T}_w^\text{ENU}$ must be set before fusion of raw GNSS measurements. We adopt a simplified version of batch initialization similar to that of \cite{GVINS}. Readers who are interested may refer to our InGVIO code for details. The scalar yaw offset between the local world frame and the true north, can either be jointly estimated or not. Though we provide both options in InGVIO, assuming $\bsz{T}_w^\text{ECEF}$ being constant will not weaken the performance.

GNSS related variables $(t_\alpha, f)$ need to be added to the state before any updates. For any newly-tracked constellation system, the initial values for $(t_\alpha, f)$ are acquired by Single Point Positioning (SPP) algorithm \cite{GPSreceiver}, and then those values are inserted into the state by delayed initialization. Some $t_\alpha$ may be removed from the state if no satellite of this constellation is observed in this epoch. 

The Jacobians of the nonlinear GNSS measurement (\ref{gvio_gnss_meas}) in the right-invariant error form are shown in (\ref{ingvio_gnss_jacobian}).
\begin{gather}
\label{ingvio_gnss_jacobian}
\begin{split}
\delta\rho^{(j)}=\bs{n}_j^T\bsz{R}_w^\text{ECEF}{^w}\hat{\bs{p}}_i^\times\delta\bs{\theta}_i^w-\bs{n}_j^T\bsz{R}_w^\text{ECEF}\delta{^w}\bs{p}_i+c\delta t_\alpha \\
\delta\dot{\rho}^{(j)}=\bs{n}_j^T\bsz{R}_w^\text{ECEF}{^w}\hat{\bs{v}}_i^\times\delta\bs{\theta}_i^w-\bs{n}_j^T\bsz{R}_w^\text{ECEF}\delta{^w}\bs{v}_i+c\delta f
\end{split}
\end{gather}
Again, $\delta\rho^{(j)}$ and $\delta\dot{\rho}^{(j)}$ are the error of pseudo range and Doppler shift to the $j$-th satellite. $\bs{n}_j$ is the unit vector pointing from the IMU to the corresponding satellite. The measurement noises weighted by satellite elevation angles are similar to those in \cite{GVINS}. Though $\chi^2$-test to avoid polluted GNSS measurements due to multi-path effect is supported in our open-sourced code, it's disabled in the experiment section to avoid biasing the evaluation of consistency maintenance.

\section{Public Dataset Experiments}\label{sec::public_datasets}

InGVIO is compared to GVINS \cite{GVINS}, an open-source state-of-the-art optimization-based algorithm, on three publicly-available datasets\footnote{https://github.com/HKUST-Aerial-Robotics/GVINS-Dataset} provided by \cite{GVINS}.
\begin{figure}[tpb]
  \centering
  \parbox{3in}{
      \centering
  \includegraphics[scale=0.52]{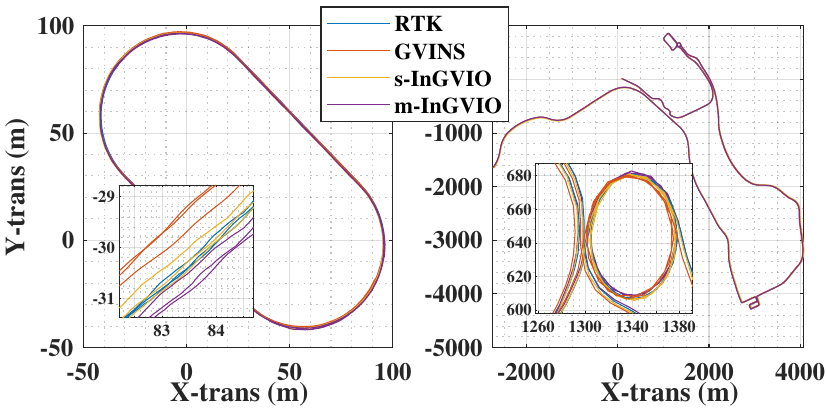}
  }
  \caption{Trajectories of InGVIO and GVINS in the $\mathtt{sports\_field}$ and $\mathtt{urban\_driving}$ datasets. The `m' and `s' implies monocular and stereo camera. RTK is viewed as the ground truth in open areas.}
  \label{fig::sf_compare}
\end{figure}
\begin{figure}[tpb]
  \centering
  \parbox{3in}{
      \centering
  \includegraphics[scale=0.55]{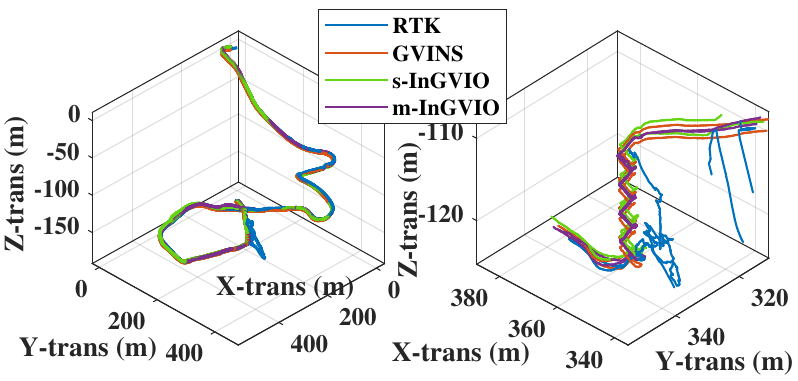}
  }
  \caption{Three-dimensional trajectories of both m/s-InGVIO and GVINS in the $\mathtt{complex\_environment}$ dataset. The left subplot gives the whole view and the right one gives local zoom when the device enters a staircase and encounters denied-GNSS situation. InGVIO and GVINS give smooth transitions between indoor and outdoor environments.}
  \label{fig::ce_ub_compare}
\end{figure}
The planar trajectories of InGVIO and GVINS tested in $\mathtt{sports\_field}$ and $\mathtt{urban\_driving}$ are shown in Fig. \ref{fig::sf_compare}. With GNSS involved, yaw and translations no longer drift. The InGVIO and GVINS estimations are much smoother than pure SPP results and fit RTK ground truth quite well.

The other dataset, namely $\mathtt{complex\_environment}$, is much more challenging due to intermittent GNSS situations. It's recorded by a handheld device traversing indoors and outdoors, and thus the receiver may lose track of all satellites and then recapture some of them later. When the receiver is near obstacles or stays indoors, the number of available satellites can be below four or even zero, when the GNSS alone cannot conduct localization. Both GVINS and InGVIO can give consistent and smooth 3-dimensional trajectories in such intermittent GNSS environments as shown in Fig. \ref{fig::ce_ub_compare}.

\begin{figure}[tpb]
  \centering
  \parbox{3in}{
      \centering
  \includegraphics[scale=0.6]{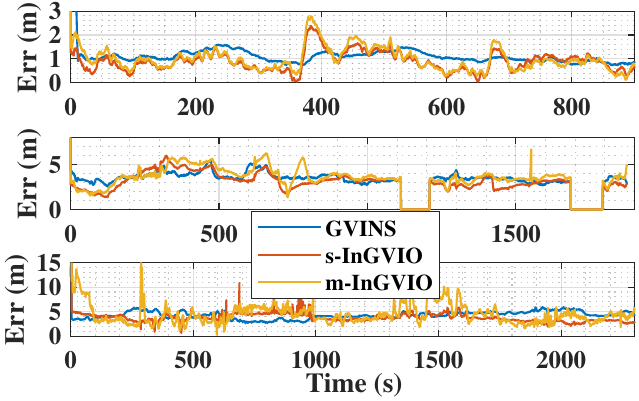}
  }
  \caption{Translational errors of InGVIO and GVINS in $\mathtt{sports\_field}$ (top), $\mathtt{complex\_environment}$ (middle) and $\mathtt{urban\_driving}$ (bottom) datasets. The error is set to zero if RTK itself is lost.}
  \label{fig::err_compare}
\end{figure}

We compare our filter-based InGVIO to optimization-based GVINS in terms of computational load and accuracy on a PC with Intel i7-12700 and 32 GB of memory. InGVIO takes the best trade-off parameter tuning in Table \ref{tab::compare_marg}, and all other conditions are kept the same, e.g. the maximal number of tracked features per frame. The accuracy of the estimation is measured by position RMSE. RTK offers the ground truth when there's an enough number of satellites in open areas. The time evolutions of the translational errors are shown in Fig. \ref{fig::err_compare}. Note that the error value is set to zero if RTK is not believable when the GNSS antenna is sheltered.

\begin{table}[tpb]
  \centering
  \caption{Comparison on GVINS Public Datasets}
  \begin{tabular}{ccccc}
  \toprule
  Dataset & Alg & Avg-Pts & RMSE (m) & Time (ms) \\
  \midrule
  $\mathtt{sp\_fi}$ & m/s-InGVIO & 140.37/122.88 & 1.08/0.98 & 2.65/3.74 \\
  $\mathtt{sp\_fi}$ & m-GVINS & 140.26 & 1.10 & 15.13 \\
  $\mathtt{co\_en}$ & m/s-InGVIO & 129.55/108.77 & 3.83/3.45 & 2.53/3.54 \\
  $\mathtt{co\_en}$ & m-GVINS & 129.75 & 3.58 & 14.67 \\
  $\mathtt{ur\_dr}$ & m/s-InGVIO & 114.99/83.46 & 4.52/3.82 & 1.86/2.14 \\
  $\mathtt{ur\_dr}$ & m-GVINS & 115.20 & 4.32 & 13.28 \\
  \bottomrule    
  \end{tabular}
  \begin{tablenotes}
      \footnotesize
      \item $\mathtt{sp\_fi}$, $\mathtt{co\_en}$ and $\mathtt{ur\_dr}$ are short for GVINS datasets $\mathtt{sports\_field}$, $\mathtt{complex\_environment}$ and $\mathtt{urban\_driving}$. The `m' and `s' indicates monocular and stereo camera. For all algorithms, the maximum number of features tracked per frame is 150 while the actual numbers are shown below `Avg-Pts'. For InGVIO, 30 cloned poses are kept in the filter.
  \end{tablenotes}
  \label{tab::compare_gvins}
\end{table}

Statistics for accuracy and speed comparison among the above algorithms are shown in Table \ref{tab::compare_gvins}. The actual average number of features per frame processed by the fusion back-end is shown in Table \ref{tab::compare_gvins} to prove that the computation consumption comparison between InGVIO and GVINS is not biased by the feature tracking front-end, since the numbers of features in monocular configuration are very close between these two algorithms. Besides, it's natural to see that the average number of features in stereo camera is less than monocular case under the same condition since failures during feature matching between the left-right camera reject some points. In monocular case, InGVIO has slightly worse accuracy (RMSE+5\%) than GVINS but just consumes less than 1/5 of the average time per frame than GVINS. In stereo case, InGVIO performs better in both accuracy and speed. This shows the obvious superiority of the efficiency of our InGVIO compared to optimization-based algorithms. In VIO, the filter-based framework sacrifices substantial accuracy for exchange of some reduction in computational load. However, with one more sensor (GNSS) and intrinsic consistency maintenance, the InGVIO filter is very competitive to current graph-based algorithms without this painful balance.

\section{Fixed-Wing Real-World Experiments}\label{sec::fixed_wing}

Apart from comparisons on public datasets, we also propose testings of InGVIO on a fixed-wing aircraft. In contrast to various well-known datasets for VIO, there are only a few publicly-available datasets containing raw GNSS data along with visual inertial measurements. The triple sensor datasets recorded by a fixed-wing aircraft are useful supplements to the GVINS datasets \cite{GVINS} which are recorded by either handheld or ground-vehicle-attached devices. Our fixed-wing datasets are open-sourced to the community.

\begin{figure}[tpb]
   \centering
   \parbox{3in}{
       \centering
   \includegraphics[scale=0.32]{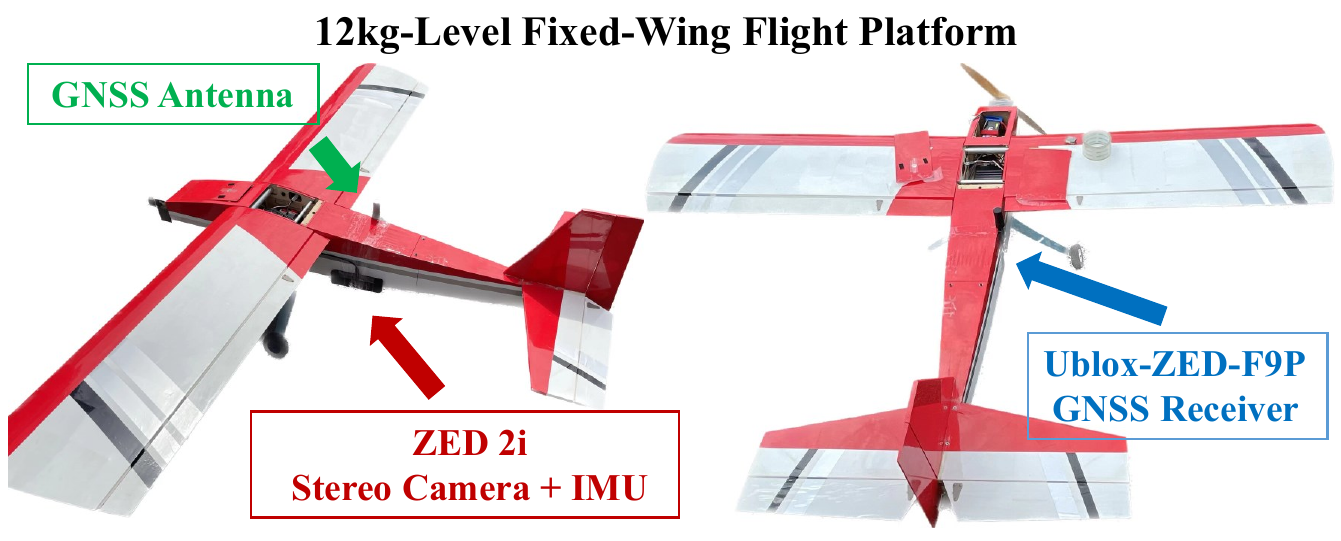}
  }
   \caption{Our fixed-wing aircraft with onboard sensors. Stereo images, IMU measurements and raw GNSS data are recorded in a TF card during flight for later evaluations.}
   \label{fig::fw_platform}
\end{figure}

The 12kg-level fixed-wing aircraft for data collection is illustrated in Fig. \ref{fig::fw_platform}. We adopt ZED 2i\footnote{https://www.stereolabs.com/zed-2i/} as our synchronized visual-inertial sensor and Ublox-ZED-F9P\footnote{https://www.u-blox.com/en/product/zed-f9p-module} as the GNSS module. RTCM messages are broadcasted from a ground station in our flight field, and then Real-Time Kinematic is enabled for ground truth by receiving such correction messages through the radio. The fixed-wing aircraft is trimmed with a head-up pitch in most flight phases, and therefore the camera is pointing downwards to maintain enough number of tracked features during climbing, cruising and descending.

\begin{figure*}[t]
  \centering
  \parbox{7in}{
      \centering
  \includegraphics[scale=0.8]{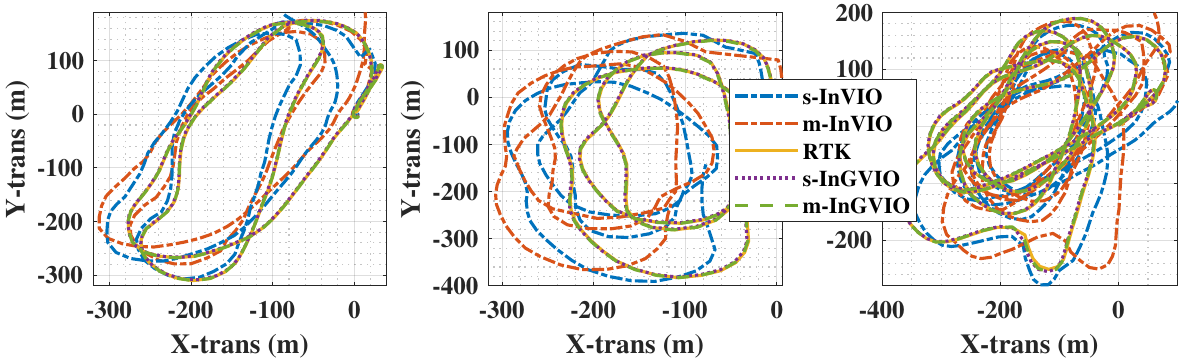}
 }
  \caption{Comparisons among invariant VIO (InVIO, Num-Sat=NONE), InGVIO and RTK ground truth in the easy- (left), medium- (middle) and hard- (right) level fixed-wing datasets. Similarly, `m/s' indicates the monocular/stereo configuration. The path lengths are 2485.3, 2911.2, 5679.3 meters, respectively.}
  \label{fig::fw}
\end{figure*}
\begin{table*}[tpb]
  \centering
  \caption{Fixed-wing Experiments on Our Proposed Datasets @ Intel i7-12700 with 32 GB Memory}
  \begin{tabular}{ccccccccc}
  \toprule
  Dataset & Num-Cam & Num-Sat & Max-KFs & Max-Pts & Avg-Pts & RMSE-Att (deg) & RMSE-Pos (m) & Avg-Time-Frame (ms) \\
  \midrule
  $\mathtt{fw\_gvi\_easy}$ & M/S & FULL & 30 & 150 & 116.63/104.35 & 1.03/0.97 & 3.18/3.16 & 2.72/3.91\\
  $\mathtt{fw\_gvi\_easy}$ & M/S & 3 & 30 & 150 & 117.89/103.51 & 1.14/1.05 & 4.26/4.19 & 2.64/3.26\\
  $\mathtt{fw\_gvi\_easy}$ & M/S & 2 & 30 & 150 & 116.64/103.87 & 4.36/4.29 & 25.78/22.64 & 2.64/3.25\\
  $\mathtt{fw\_gvi\_easy}$ & M/S & NONE & 30 & 150 & 117.21/103.68 & 7.88/7.65 & 34.24/30.11 & 2.63/3.24 \\   
  $\mathtt{fw\_gvi\_medium}$ & M/S & FULL & 30 & 150 & 122.83/103.64 & 1.04/0.99 & 3.27/2.94 & 2.89/3.57 \\
  $\mathtt{fw\_gvi\_medium}$ & M/S & 3 & 30 & 150 & 121.94/103.47 & 1.30/1.19 & 6.63/6.23 & 2.78/3.42 \\
  $\mathtt{fw\_gvi\_medium}$ & M/S & 2 & 30 & 150 & 122.46/103.38 & 8.01/7.83 & 64.77/40.19 & 2.79/3.42 \\
  $\mathtt{fw\_gvi\_medium}$ & M/S & NONE & 30 & 150 & 122.88/103.41 & 25.92/22.78 & 82.13/54.55 & 2.77/3.40 \\
  $\mathtt{fw\_gvi\_hard}$ & M/S & FULL & 30 & 150 & 124.57/104.54 & 1.07/0.99 & 3.43/3.24 & 2.84/4.22 \\
  $\mathtt{fw\_gvi\_hard}$ & M/S & 3 & 30 & 150 & 124.03/103.95 & 1.33/1.27 & 6.95/6.10 & 2.46/3.66 \\
  $\mathtt{fw\_gvi\_hard}$ & M/S & 2 & 30 & 150 & 123.63/104.13 & 10.25/8.94 & 67.61/45.99 & 2.43/3.66 \\
  $\mathtt{fw\_gvi\_hard}$ & M/S & NONE & 30 & 150 & 124.30/103.32 & 26.49/22.32 & 78.62/51.77 & 2.42/3.63 \\
  \bottomrule    
  \end{tabular}
  \begin{tablenotes}
    \footnotesize
    \item[*] `M' is short for monocular camera and `S' is stereo camera. `KF' indicates the maximum number of cloned poses kept in the filter. `Max-Pts' is the preset maximal number of features tracked in each frame. `Avg-Pts' is the actual average features in each frame. Attitude and position RMSEs are calculated via the ground truth generated by an EKF combining IMU and RTK. The average processing time only involves the fusion back-end.
  \end{tablenotes}
  \label{tab::compare_fw}
\end{table*}

The down-pointing camera greatly raises the difficulties for initialization of the VIO. Our fixed-wing aircraft remains stationary at the beginning of each dataset and then quickly starts takeoff run. The distance from the camera to the runway is very close and the runway pavement lacks stable features for the front-end. This will cause lots of trouble for the VIO until the aircraft gains enough height, not to mention the violent acceleration during initial roll and climb. Thanks to the good convergence properties of the invariant filter framework, our InGVIO goes smoothly in all fixed-wing datasets with or without GNSS. To the best of our tuning, the VIO part of GVINS cannot be successfully initialized no matter on the ground or in the air and thus the fusion of GNSS in GVINS cannot be further initiated. GVINS is not included in the comparisons in this section.

Three datasets are proposed herein with different difficulty levels. Difficulty is evaluated from the aspect of trajectory shapes, flight duration and maneuvers. The estimated flight paths of the fixed-wing datasets are illustrated in Fig. \ref{fig::fw}. Unlike the traditional `Airfield Traffic Pattern' in the easy-level dataset, $\mathtt{fw\_gvi\_medium}$ contains continuous spirals and one more loop. The most challenging flight-test is $\mathtt{fw\_gvi\_hard}$, where a great many steep turns are conducted and the height of the aircraft varies from 20 to 80 meters. We tried our best to operate our fixed-wing aircraft to reach its vertical speed limits and the airspeed reaches a maximum value of 25 m/s. The fixed-wing experiments reveal the fact that InGVIO overcomes the drifting of unobservable directions and is capable of navigating in global frames.

Unlike slow motions in the GVINS datasets, using EKF to fuse RTK measurements and IMU on the fast-flying fixed-wing can give reliable 6-DoF ground truth. Speed and accuracy of these flight-tests are analyzed in Table \ref{tab::compare_fw}. Besides, low satellite number experiments are conducted to show the benefits of intrinsic consistency preserving in InGVIO. Since the vehicle is operated in open areas, no potential multi-path effect happens and thus $\chi^2$-tests are disabled to avoid affecting the validation of the algorithm. The results show that accuracy slightly degrades in 3-satellite case and drifting is obviously smaller than pure VIO in 2-satellite case.

The meter-level difference of the estimated path of GVINS\cite{GVINS} or InGVIO to the RTK is attributed to the inaccuracy of the atmospherical propagation models of the pseudo range. GVINS and InGVIO will gradually converge to SPP results when the number of satellites is enough. By respecting the geometry, even if in open areas when the system is fully observable, InGVIO is preferred over conventional methods due to its IMU propagation with no linearization error \cite{InEKF-Stable-Observer}.

\section{Conclusions}\label{sec:c}

In this letter, an invariant filter approach InGVIO to fuse visual, inertial and raw GNSS is proposed and open-sourced to the community. We demonstrate that the nonlinear log-error of the state in matrix Lie group form automatically captures all presented infinitesimal symmetry structures of the GVIO system even in degenerate motions, and therefore InGVIO is intrinsically consistent. Moreover, a powerful albeit concise marginalization strategy is proposed to extend the baseline in feature triangulation and meanwhile control the computational consumption of visual updates and anchor change. It's a trade-off between speed and accuracy making MSCKF features dominate the visual updates in contrast to \cite{OpenVINS,Decoupled-RIEKF} and thus InGVIO is superiorly fast. InGVIO is tested and compared to the optimization-based GVINS \cite{GVINS} on open datasets to show it reduces computational consumption while possessing the same or even better accuracy level over the former. Besides, we propose open-sourced datasets containing visual, inertial and raw GNSS data with ground truth on our fixed-wing platform, which are strong supplements to the current GVINS datasets. InGVIO gives smooth and small-drifting pose estimations in the above fixed-wing datasets among all difficulty levels with various satellite numbers. 




\section*{APPENDIX}

Auxiliary $\Gamma$-functions $\mathbb{R}^3\mapsto\mathbb{R}^{3\times 3}$ are defined by power series $\Gamma_m(\bs{\theta})=\sum_{n=0}^{+\infty}\frac{(\bs{\theta}^\times)^n}{(m+n)!}$. $\Gamma_m$ has explicit analytical expression \cite{ContactAid}. Specifically, we have $\exp(\bs{\theta}^\times)=\Gamma_0(\bs{\theta})$ and $J_l(\bs{\theta})=\Gamma_1(\bs{\theta})$, where $J_l$ is the left Jacobian of $\mathbb{SO}(3)$.

\section*{ACKNOWLEDGMENT}

The authors would like to thank Qianbo Xiao, Tianheng Yao and Pengfei Zhang for assisting fixed-wing operation.

\bibliographystyle{IEEEtran.bst}
\bibliography{ingvio.bib}

\begin{thebibliography}{10}
\providecommand{\url}[1]{#1}
\csname url@samestyle\endcsname
\providecommand{\newblock}{\relax}
\providecommand{\bibinfo}[2]{#2}
\providecommand{\BIBentrySTDinterwordspacing}{\spaceskip=0pt\relax}
\providecommand{\BIBentryALTinterwordstretchfactor}{4}
\providecommand{\BIBentryALTinterwordspacing}{\spaceskip=\fontdimen2\font plus
\BIBentryALTinterwordstretchfactor\fontdimen3\font minus
  \fontdimen4\font\relax}
\providecommand{\BIBforeignlanguage}[2]{{%
\expandafter\ifx\csname l@#1\endcsname\relax
\typeout{** WARNING: IEEEtran.bst: No hyphenation pattern has been}%
\typeout{** loaded for the language `#1'. Using the pattern for}%
\typeout{** the default language instead.}%
\else
\language=\csname l@#1\endcsname
\fi
#2}}
\providecommand{\BIBdecl}{\relax}
\BIBdecl

\bibitem{MSCKF1.0}
A.~I. Mourikis and S.~I. Roumeliotis, ``A {{Multi}}-{{State Constraint Kalman
  Filter}} for {{Vision}}-aided {{Inertial Navigation}},'' in \emph{Proceedings
  2007 {{IEEE International Conference}} on {{Robotics}} and
  {{Automation}}}.\hskip 1em plus 0.5em minus 0.4em\relax {Rome, Italy}:
  {IEEE}, Apr. 2007, pp. 3565--3572.

\bibitem{FEJ1.0}
M.~Li and A.~I. Mourikis, ``Improving the accuracy of {{EKF}}-based
  visual-inertial odometry,'' in \emph{2012 {{IEEE International Conference}}
  on {{Robotics}} and {{Automation}}}.\hskip 1em plus 0.5em minus 0.4em\relax
  {St Paul, MN, USA}: {IEEE}, May 2012, pp. 828--835.

\bibitem{FEJ2.0}
C.~Chen, Y.~Yang, P.~Geneva, and G.~Huang, ``Fej2: A consistent visual-inertial
  state estimator design,'' in \emph{2022 International Conference on Robotics
  and Automation (ICRA)}, 2022, pp. 9506--9512.

\bibitem{OC-VINS}
J.~A. Hesch, D.~G. Kottas, S.~L. Bowman, and S.~I. Roumeliotis, ``Consistency
  analysis and improvement of vision-aided inertial navigation,'' \emph{IEEE
  Transactions on Robotics}, vol.~30, no.~1, pp. 158--176, 2014.

\bibitem{ROVIO}
M.~Bloesch, M.~Burri, S.~Omari, M.~Hutter, and R.~Siegwart, ``Iterated extended
  {{Kalman}} filter based visual-inertial odometry using direct photometric
  feedback,'' \emph{The International Journal of Robotics Research}, vol.~36,
  no.~10, pp. 1053--1072, Sep. 2017.

\bibitem{MSCKF-Hover}
D.~G. Kottas, K.~J. Wu, and S.~I. Roumeliotis, ``Detecting and dealing with
  hovering maneuvers in vision-aided inertial navigation systems,'' in
  \emph{2013 {{IEEE}}/{{RSJ International Conference}} on {{Intelligent
  Robots}} and {{Systems}}}.\hskip 1em plus 0.5em minus 0.4em\relax {Tokyo}:
  {IEEE}, Nov. 2013, pp. 3172--3179.

\bibitem{Robocentric-VIO}
Z.~Huai and G.~Huang, ``Robocentric visual-inertial odometry,'' in \emph{2018
  IEEE/RSJ International Conference on Intelligent Robots and Systems (IROS)},
  2018, pp. 6319--6326.

\bibitem{Exploiting-Symmetries-SLAM}
M.~Brossard, A.~Barrau, and S.~Bonnabel, ``Exploiting symmetries to design ekfs
  with consistency properties for navigation and slam,'' \emph{IEEE Sensors
  Journal}, vol.~19, no.~4, pp. 1572--1579, 2019.

\bibitem{OpenVINS}
P.~Geneva, K.~Eckenhoff, W.~Lee, Y.~Yang, and G.~Huang, ``{{OpenVINS}}: A
  {{Research Platform}} for {{Visual}}-{{Inertial Estimation}},'' in \emph{2020
  {{IEEE International Conference}} on {{Robotics}} and {{Automation}}
  ({{ICRA}})}.\hskip 1em plus 0.5em minus 0.4em\relax {Paris, France}: {IEEE},
  May 2020, pp. 4666--4672.

\bibitem{RI-MSCKF}
K.~Wu, T.~Zhang, D.~Su, S.~Huang, and G.~Dissanayake, ``An invariant-ekf vins
  algorithm for improving consistency,'' in \emph{2017 IEEE/RSJ International
  Conference on Intelligent Robots and Systems (IROS)}, 2017, pp. 1578--1585.

\bibitem{Decoupled-RIEKF}
Y.~Yang, C.~Chen, W.~Lee, and G.~Huang, ``Decoupled right invariant error
  states for consistent visual-inertial navigation,'' \emph{IEEE Robotics and
  Automation Letters}, vol.~7, no.~2, pp. 1627--1634, 2022.

\bibitem{EqF-VIO}
P.~v. Goor and R.~Mahony, ``An equivariant filter for visual inertial
  odometry,'' in \emph{2021 IEEE International Conference on Robotics and
  Automation (ICRA)}, 2021, pp. 14\,432--14\,438.

\bibitem{VINS-Concise-Review}
G.~Huang, ``Visual-inertial navigation: A concise review,'' in \emph{2019
  International Conference on Robotics and Automation (ICRA)}, 2019, pp.
  9572--9582.

\bibitem{S-MSCKF}
K.~Sun, K.~Mohta, B.~Pfrommer, M.~Watterson, S.~Liu, Y.~Mulgaonkar, C.~J.
  Taylor, and V.~Kumar, ``Robust stereo visual inertial odometry for fast
  autonomous flight,'' \emph{IEEE Robotics and Automation Letters}, vol.~3,
  no.~2, pp. 965--972, 2018.

\bibitem{InEKF-Stable-Observer}
A.~Barrau and S.~Bonnabel, ``The invariant extended kalman filter as a stable
  observer,'' \emph{IEEE Transactions on Automatic Control}, vol.~62, no.~4,
  pp. 1797--1812, 2017.

\bibitem{Consistent-EKF-SLAM}
\BIBentryALTinterwordspacing
------. (2015) An ekf-slam algorithm with consistency properties. [Online].
  Available: \url{https://arxiv.org/pdf/1510.06263}
\BIBentrySTDinterwordspacing

\bibitem{CCAnalysis-RI-MSCKF}
T.~Zhang, K.~Wu, J.~Song, S.~Huang, and G.~Dissanayake, ``Convergence and
  consistency analysis for a 3-d invariant-ekf slam,'' \emph{IEEE Robotics and
  Automation Letters}, vol.~2, no.~2, pp. 733--740, 2017.

\bibitem{Equi-Filter}
P.~van Goor, T.~Hamel, and R.~Mahony, ``Equivariant filter (eqf),'' \emph{IEEE
  Transactions on Automatic Control}, pp. 1--13, 2022.

\bibitem{EqF-CDC}
------, ``Equivariant filter (eqf): A general filter design for systems on
  homogeneous spaces,'' in \emph{2020 59th IEEE Conference on Decision and
  Control (CDC)}, 2020, pp. 5401--5408.

\bibitem{EqVIO}
\BIBentryALTinterwordspacing
P.~van Goor and R.~Mahony. (2022) Eqvio: An equivariant filter for visual
  inertial odometry. [Online]. Available:
  \url{https://doi.org/10.48550/arXiv.2205.01980}
\BIBentrySTDinterwordspacing

\bibitem{AdaptiveFusionGVIO}
Z.~Gong, P.~Liu, F.~Wen, R.~Ying, X.~Ji, R.~Miao, and W.~Xue, ``Graph-based
  adaptive fusion of gnss and vio under intermittent gnss-degraded
  environment,'' \emph{IEEE Transactions on Instrumentation and Measurement},
  vol.~70, pp. 1--16, 2021.

\bibitem{LooseCoupleGVIO}
G.~Cioffi and D.~Scaramuzza, ``Tightly-coupled {{Fusion}} of {{Global
  Positional Measurements}} in {{Optimization}}-based {{Visual}}-{{Inertial
  Odometry}},'' in \emph{2020 {{IEEE}}/{{RSJ International Conference}} on
  {{Intelligent Robots}} and {{Systems}} ({{IROS}})}.\hskip 1em plus 0.5em
  minus 0.4em\relax {Las Vegas, NV, USA}: {IEEE}, Oct. 2020, pp. 5089--5095.

\bibitem{GVINS}
S.~Cao, X.~Lu, and S.~Shen, ``Gvins: Tightly coupled gnss–visual–inertial
  fusion for smooth and consistent state estimation,'' \emph{IEEE Transactions
  on Robotics}, vol.~38, no.~4, pp. 2004--2021, 2022.

\bibitem{OB-GVINS}
J.~Liu, W.~Gao, and Z.~Hu, ``Optimization-based visual-inertial slam tightly
  coupled with raw gnss measurements,'' in \emph{2021 IEEE International
  Conference on Robotics and Automation (ICRA)}, 2021, pp. 11\,612--11\,618.

\bibitem{P3-GVINS}
T.~Li, L.~Pei, Y.~Xiang, W.~Yu, and T.-K. Truong, ``P$^{3}$-vins:
  Tightly-coupled ppp/ins/visual slam based on optimization approach,''
  \emph{IEEE Robotics and Automation Letters}, vol.~7, no.~3, pp. 7021--7027,
  2022.

\bibitem{VOC-GSMSCKF}
C.~Liu, C.~Jiang, and H.~Wang, ``Variable observability constrained
  visual-inertial-gnss ekf-based navigation,'' \emph{IEEE Robotics and
  Automation Letters}, vol.~7, no.~3, pp. 6677--6684, 2022.

\bibitem{GAINS}
W.~Lee, P.~Geneva, Y.~Yang, and G.~Huang, ``Tightly-coupled gnss-aided
  visual-inertial localization,'' in \emph{2022 International Conference on
  Robotics and Automation (ICRA)}, 2022, pp. 9484--9491.

\bibitem{ConAnalysisVINS}
G.~Huang, M.~Kaess, and J.~J. Leonard, ``Towards consistent visual-inertial
  navigation,'' in \emph{2014 {{IEEE International Conference}} on {{Robotics}}
  and {{Automation}} ({{ICRA}})}.\hskip 1em plus 0.5em minus 0.4em\relax {Hong
  Kong, China}: {IEEE}, May 2014, pp. 4926--4933.

\bibitem{GPSreceiver}
J.~B.-y. Tsui, \emph{Fundamentals of Global Positioning System Receivers a
  Software Approach}, 2nd~ed.\hskip 1em plus 0.5em minus 0.4em\relax {New
  Jersey}: {New Jersey : John Wiley \& Sons, Inc.}, 2005.

\bibitem{IonoDelay}
J.~A. Klobuchar, ``Ionospheric time-delay algorithm for single-frequency gps
  users,'' \emph{IEEE Transactions on Aerospace and Electronic Systems}, vol.
  AES-23, no.~3, pp. 325--331, 1987.

\bibitem{TropoDelay}
J.~Saastamoinen, ``Contributions to the theory of atmospheric refraction,''
  \emph{Bulletin g\'eod\'esique}, vol. 105, no.~1, pp. 279--298, 1972.

\bibitem{SagnacEffect}
N.~Ashby, ``Relativity in the {{Global Positioning System}},'' \emph{Living
  Reviews in Relativity}, vol.~6, no.~1, p.~1, Jan. 2003.

\bibitem{ContactAid}
R.~Hartley, M.~Ghaffari, R.~M. Eustice, and J.~W. Grizzle, ``Contact-aided
  invariant extended kalman filtering for robot state estimation,'' \emph{The
  International Journal of Robotics Research}, vol.~39, no.~4, pp. 402--430,
  2020.

\end{thebibliography}

\end{document}